\documentclass[letterpaper, 10 pt, conference]{ieeeconf}  % Comment this line out if you need a4paper

\IEEEoverridecommandlockouts                              % This command is only needed if 
\usepackage{amsmath} % assumes amsmath package installed

\title{\LARGE \bf
Policy-Induced Hand Priors in Humanoid Dual-Arm Manipulation: Diagnosing and Mitigating Initial-Pose Dependence
}

\author{Chaeyeon Jung and Juyoun Park$^{*}$%
\thanks{$^{*}$Corresponding author.}%
\thanks{Center for Humanoid Research, Korea Institute of Science and Technology (KIST), Republic of Korea.}%
}

\usepackage{cite}
\usepackage{booktabs}
\usepackage{graphicx}
\usepackage{array}
\usepackage{makecell}
\usepackage{placeins}
\usepackage{float}
\usepackage{xurl}

\begin{document}

\maketitle
\thispagestyle{empty}
\pagestyle{empty}

%%%%%%%%%%%%%%%%%%%%%%%%%%%%%%%%%%%%%%%%%%%%%%%%%%%%%%%%%%%%%%%%%%%%%%%%%%%%%%%%
\begin{abstract}

Vision-language-action (VLA) policies are expected to operate robustly across variations in the robot's initial configuration, yet aggregate task success can conceal pose-specific failures and inappropriate hand selection. This work investigates initial-pose dependence in VLA-based humanoid dual-arm manipulation. We characterize the initial-condition-dependent early hand preference  as a policy-induced hand prior and quantify it using HandPriorScore, residual hand bias, and target responsiveness. Evaluations across multiple policies and 17 initial configurations reveal strong initial-pose--policy interactions: the same pose produces substantially different success rates across policies, while a single policy exhibits large performance variation across poses. Specific initial arm configurations can suppress or induce an asymmetric hand preference, with the resulting effect varying in direction and strength across policies. Wrist-camera observations also influence hand selection and task performance. Expanding initial-pose coverage in the training dataset substantially improves robustness, while targeted augmentation around a low-performing configuration increases its success rate. Comparisons across training configurations show that sufficient exposure to the target simulation task is beneficial, whereas the effect of real or auxiliary data depends on pose coverage, simulation ratio, and observation availability. These findings characterize a pose-conditioned hand prior, identify a localized initial arm configuration as a causal handle on hand-selection behavior, and demonstrate how data coverage and training composition affect initial-pose robustness.

\end{abstract}

%%%%%%%%%%%%%%%%%%%%%%%%%%%%%%%%%%%%%%%%%%%%%%%%%%%%%%%%%%%%%%%%%%%%%%%%%%%%%%%%
\section{INTRODUCTION}

Recent robot policies and vision-language-action (VLA) models aim to learn generalist robot behaviors from multimodal observations and large-scale robot demonstrations \cite{RT1,RT-2,OXE,Octo,OpenVLA,GR00T}. Such policies are expected to operate robustly across task conditions, including variations in the robot's initial configuration. However, several evaluation protocols in humanoid control and robot manipulation use reference-aligned or otherwise fixed initializations. Individual humanoid skills and motion-tracking tasks have been evaluated from the first pose of a reference or target motion \cite{learning_to_sit,meta_motivo}, while Mimicking-Bench initializes the humanoid from the first frame of a sampled interaction sequence \cite{mimicking_bench}. In manipulation, OpenVLA performs paired evaluations using matched initial robot and object states and fixes the end-effector starting configuration across rollouts for most tasks \cite{OpenVLA}. These choices facilitate controlled and reproducible comparison, but they can make measured performance conditional on the selected initial configuration.

This concern is not merely procedural. Recent robustness studies have shown that modest changes in camera viewpoints and robot initial states can cause severe degradation in VLA performance, even for policies that achieve near-perfect success under standard benchmark conditions \cite{libero_plus}. However, an initial-pose shift is inherently multimodal: it simultaneously changes the robot’s proprioceptive state, its camera-visible body configuration, and when available, the hand–-object geometry observed from wrist-mounted cameras. Consequently, performance degradation alone cannot determine whether the failure originates from the proprioceptive state, visual self-pose cues, or their interaction. Recent analyses of vision-proprioception policies further suggest that policies may favor concise proprioceptive signals during training, suppressing the use of visual evidence when visual localization becomes necessary \cite{gap}. Related studies further report that policies can overfit to proprioceptive trajectories or continue action progression despite mismatched visual evidence \cite{state_free,revip}. However, it remains unclear whether this imbalance can cause an initial robot configuration to act as a shortcut for hand selection in humanoid dual-arm manipulation.

Aggregate task success can further obscure such initial-pose-dependent failures. A policy may achieve high average performance while consistently failing in a particular pose or selecting an inappropriate hand under a specific initial configuration. In our GR00T-based Unitree G1 setting \cite{GR00T,GR00T_N15, unitree_g1, unitree_dex3, unitree_dex3_datasets}, we observe that identical object configurations can induce different hand-selection behaviors depending on the robot’s initial pose. In particular, a specific initial configuration repeatedly triggers right-hand preferences even when the target object is located on the left, but nearby initial configurations in the initial-state space of a training dataset produce more object-consistent hand choices. This behavior is not merely a deviation in the approach trajectory, but a change in the action mode selected by the policy. We refer to this empirical, initial-condition-dependent preference as \textbf{a policy-induced hand prior}. 

This observation raises three questions. First, can localized changes in the initial arm configuration alter the observed hand prior? Second, how do wrist-camera observations influence hand preference and task performance? Third, how do initial-pose coverage, camera observations, and training-data composition affect robustness across initial configurations?

To determine how initial-pose cues influence hand selection, we combine physically consistent closed-loop joint interventions with paired wrist-camera masking analyses. In simulation, we vary the initial configuration of the Unitree G1 while controlling the object placement and evaluate the resulting changes in hand selection and task performance. We further perform a bidirectional hybrid intervention by exchanging only selected arm joint values between two behaviorally distinct initial poses, testing whether a localized change in the physical state can switch the policy’s hand-selection behavior. To quantify the resulting early hand preference, we use HandPriorScore, a continuous normalized measure of left--right action imbalance computed before grasp. We further compare intermediate visual and action representations, using state-relative action changes $(\Delta a_t=a_t-q_t)$ to separate commanded motion from the shared offset of absolute joint targets. Separately, we pair each full observation with left-, right-, and both-wrist-camera-masked inputs from the same model, condition, object location, and episode. Because modality masking can generate out-of-distribution inputs, we interpret these ablations as measures of input-channel sensitivity rather than causal contribution percentages. Finally, we add training poses near a weak evaluation configuration to examine when local initial-pose coverage improves generalization.

Together, these experiments yield two main contributions. First, we characterize initial-pose dependence in humanoid dual-arm manipulation as a policy-induced hand prior and quantify its early directional behavior using HandPriorScore, residual hand bias, and target responsiveness. Wrist-camera masking shows that wrist observations influence hand preference and task performance but do not independently explain the wrong-hand behavior, while bidirectional joint-level interventions identify the local configuration of $R_{\mathrm{arm},3}$ and $R_{\mathrm{arm},5}$ as a causal handle on the pose-conditioned hand prior. Second, we analyze how initial-pose coverage, wrist-camera observations, auxiliary-data composition, and simulation sampling affect initial-pose robustness. Expanding a target-coupled two-pose dataset to eight diversified poses substantially improves robustness, and targeted augmentation around a low-performing configuration increases its success rate from 30\% to 75\% at a previously low-performing initial pose. Comparisons across training configurations further show that wrist observations and sufficient target-task exposure are beneficial, whereas the effect of real or auxiliary data depends on pose coverage, simulation ratio, and observation availability.

%%%%%%%%%%%%%%%%%%%%%%%%%%%%%%%%%%%%%%%%%%%%%%%%%%%%%%%%%%%%%%%%%%%%%%%%%%%%%%%%

\section{RELATED WORK}

\subsection{VLA Policies and Robustness under Perturbations}

VLA models learn end-to-end robot policies that map multimodal inputs, including visual observations, language instructions, and proprioceptive states to action predictions. Recent works have shown that large-scale vision-language pretraining can be combined with robot demonstrations to build generalist manipulation policies. These advances have improved generalization across tasks, robot platforms, and environments, making VLA models increasingly suitable for complex manipulation and robot control. 

Benchmark suites such as LIBERO evaluate knowledge transfer and
generalization across diverse manipulation tasks \cite{LIBERO}. Recent generalist robot policies report strong task-level performance
across their respective evaluation suites \cite{Octo,OpenVLA,GR00T}.
Despite aggregate success metrics, they provide limited insight into how end-to-end policies behave under different initial robot configurations. They do not reveal whether a policy selects its actions based on object observations or instead exploits shortcuts from the initial robot state.

Building on LIBERO, LIBERO-Plus evaluates VLA robustness under controlled
perturbations to camera viewpoints, robot initial states, object layouts,
and other environmental factors \cite{libero_plus}. This is particularly relevant because the initial robot configuration is encoded not only in proprioceptive states but also in visual observations through the camera-visible pose of the robot. However, these benchmarks mainly quantify performance degradation, leaving open how initial-state perturbations translate into shortcut behaviors, such as hand-selection bias. This gap is especially important in humanoid dual-arm manipulation, where the initial robot configuration can influence both the approach trajectory and the choice of which hand to use.

\subsection{Vision-Proprioception Bias and State Shortcut Learning}

Prior studies on vision-proprioception imbalance suggest that proprioceptive state can both support and bias policy learning. When proprioceptive inputs provide compact predictive cues, policies may underuse visual evidence, overfit to training state-action trajectories, or become brittle under shifts in absolute state distributions. In particular, GAP shows that policies may favor concise proprioceptive signals during training, suppressing visual learning when visual localization becomes necessary \cite{gap}. State-free Policy similarly reports that policies can overfit to proprioceptive trajectories and exhibit poor spatial generalization, and shows that relative end-effector actions together with sufficient wrist-camera observations can reduce this dependence \cite{state_free}. Related VLA analyses have identified similar modality imbalance. ReViP attributes visually inconsistent policy behavior to state-dominant fusion, in which the policy follows its internal state progression while underusing visual evidence \cite{revip}. Separately, studies of proprioceptive representation show that absolute state encodings can be brittle under changes in the robot reference frame, whereas episode-relative representations can improve out-of-distribution robustness \cite{absolute_state}.

These findings raise the question of whether the initial robot state can act not only as control information, but also as a shortcut for selecting an action mode. In this study, we probe how initial arm configuration and wrist observations contribute to initial-pose dependence and analyze how pose coverage and training composition affect robustness.

%%%%%%%%%%%%%%%%%%%%%%%%%%%%%%%%%%%%%%%%%%%%%%%%%%%%%%%%%%%%%%%%%%%%%%%%%%%%%%%%

\section{Policy-induced hand prior}

\subsection{Problem Formulation} 

To characterize how a learned policy maps initial conditions to early hand-selection behavior, we first formalize the policy-induced hand prior. Borrowing the notion of a prior distribution, we define the policy-induced hand prior as the initial-condition-dependent preference of a learned policy over early hand-action modes.

%Let $D_{\mathrm{train}}$ denote the training-data distribution and $\mathcal{A}$ the policy-training procedure. The parameters of the policy trained on $D_{\mathrm{train}}$ are given by
%$\theta_D=\mathcal{A}(D_{\mathrm{train}})$.
%For the resulting policy $\pi_{\theta_D}$, we conceptually represent its conditional tendency over early hand-action modes as

Let $D$ denote the training-data distribution and $\theta_D$ the parameters of the policy trained on $D$. For the resulting policy $\pi_{\theta_D}$, we conceptually represent its conditional preference over early hand-action modes as

\begin{equation}
p_{\pi_{\theta_D}}
\left(
h \mid s_0,o_0,x_{\mathrm{apple}}
\right)
\label{eq:hand_prior_distribution}
\end{equation}
where $ h\in\{
\mathrm{left},
\mathrm{right},
\mathrm{bimanual},
\mathrm{undecided}
\}.$ Here, $s_0$ denotes the initial proprioceptive state, $o_0$ denotes the initial visual observation from the high and wrist cameras, and $x_{\mathrm{apple}}\in\{\mathrm{left}, \mathrm{center}, \mathrm{right}\}$ denotes the location of the target object, i.e. the apple. The variable $h$ represents the hand-action mode expressed during the initial decision window, 
\begin{equation}
\mathcal{W}_T=\{t_0,\ldots,t_0+T-1\}
\label{window}.
\end{equation}
The policy does not explicitly output the distribution in \eqref{eq:hand_prior_distribution}. Instead, we operationalize its initial directional preference using the action-based HandPriorScore defined below. %We additionally use TotalActivity to distinguish bimanual competition from low-activity or unresolved behavior.

\subsection{Initial Decision Window HandPriorScore}

We quantify the policy’s early left--right hand preference during the initial decision window using relative joint actions. The relative action of each joint is defined as the difference between the target joint position predicted by the policy and the current joint state.

Let  $S \in \{L,R\}$ denote the robot side, $G\in\{\mathrm{arm},\mathrm{hand}\}$ the joint group, and $j \in \{1,\ldots,d_S^G\}$ the joint index within that group. In our setup, both arms and both Dex3-1 hands are represented using seven joint dimensions per side \cite{unitree_g1,unitree_dex3,unitree_lerobot}, giving $d_{L}^{\mathrm{arm}} = d_{R}^{\mathrm{arm}} = d_{L}^{\mathrm{hand}} = d_{R}^{\mathrm{hand}} = 7$. The relative action of joint j at time t is defined as
\begin{equation}
\Delta a_{S,t,j}^{G} = a_{S,t,j}^{G} - q_{S,t,j}^{G}
\label{eq:relative_action}
\end{equation}
In this formulation, $a_{S,t,j}^{G}$ denotes the target joint position predicted by the policy, and $q_{S,t,j}^{G}$ denotes the current joint state at the same time step. Their difference, $\Delta a_{S,t,j}^{G}$, represents the magnitude and direction of the commanded joint motion relative to the current robot configuration. Directly aggregating relative actions with different joint-wise ranges and variances may cause high-variance joints to dominate the resulting score. To prevent this, we normalize each relative action using a shared joint-wise scale. We estimate a shared joint-wise normalization scale from the full-observation evaluation rollouts pooled across all evaluated policies:
$$
\sigma_{S,j}^{G}=
\operatorname{Std}_{e\in\mathcal{R}_{\mathrm{norm}},
\,t\in\mathcal{W}_{T}}
\left[
\Delta a_{e,S,t,j}^{G}
\right]$$
The normalization pool $\mathcal{R}_{\mathrm{norm}}$ is not a separate held-out dataset. It is used only to define a common per-joint scale across all policies and conditions. Thus, $\sigma_{S,j}^{G}$ serves as an in-sample normalizer. Applying the same normalization scale to every evaluated policy enables their HandPriorScores to be compared under a common scale.
%For the masking analyses, $\sigma_{S,j}^{G}$ is estimated using only the full-observation rollouts, and the same scale is then applied to both the full-observation and masked conditions.
Using this scale, we compute the root-mean-square magnitude of the normalized relative actions, which removes the sign of each joint action and measures the overall amount of commanded motion for robot side $S$ and joint group $G$ at time $t$:
$$M_{S,t}^{\mathrm{G}}=\sqrt{\frac{1}{d_{S}^{G}}\sum_{j=1}^{d_{S}^{G}} {\left(\frac{\Delta a_{S,t,j}^{\mathrm{G}}}{{\sigma }_{S,j}^{\mathrm{G}}+\epsilon }\right)}^{2}}$$

The small constant $\epsilon$ prevents numerical instability when the standard deviation is close to zero. Within this window in \eqref{window}, the average action activity of each side is computed by combining the normalized arm and hand action magnitudes:
$$A_{S}^{\left(T\right)}=\frac{1}{T}\sum_{t\in\mathcal{W}_T}\left(M_{S,t}^{\mathrm{arm}}+\lambda M_{S,t}^{\mathrm{hand}}\right)$$
The weight $\lambda$ controls the contribution of the hand action magnitude to the total activity of each side. The direct difference between the right- and left-side activities is defined as the raw HandPriorScore:
$$
H_{\mathrm{raw}}^{\left(T\right)}=A_{R}^{\left(T\right)}-A_{L}^{\left(T\right)}
$$
Since the raw difference is affected by the overall action magnitude, we normalize it by the sum of the left and right-side activities to obtain the HandPriorScore:
\begin{equation}
H^{\left(T\right)}=\frac{A_{R}^{\left(T\right)}-A_{L}^{\left(T\right)}}{A_{R}^{\left(T\right)}+A_{L}^{\left(T\right)}+\epsilon }
\label{eq:normalized HandPriorScore}
\end{equation}
A positive $H^{(T)}$ indicates a right-hand directional preference, whereas a negative $H^{(T)}$ indicates a left-hand directional preference. A larger $\left|H^{(T)}\right|$ indicates a stronger directional commitment toward one side. By contrast, a value close to zero may indicate either similar activity levels between the two sides or insufficient overall motion to reveal a clear directional preference.	
%To distinguish different behaviors when $H^(T)$ is close to zero, we define TotalActivity as the sum of the left- and right-side activities:
% $$ E^{(T)} = A_R^{(T)}+A_L^{(T)} \label{TotalActivity} $$
%A high $E^(T)$ together with $H^{(T)}\approx0$ suggests that both sides are simultaneously active at similar magnitudes, which we interpret as bimanual competition. In contrast, a low $E^(T)$ together with $H^{(T)}\approx0$ indicates that the policy has not yet clearly committed to either hand or that little initial motion has occurred. - Results에 정량화 하지 않은 부분이라 생략    

To separate the influence of target object location from the policy’s overall directional preference, we compute the mean HandPriorScore for each target location x as 
$$
\mu_x^{(T)}=\frac{1}{n_x}\sum_{e:x_e=x}H_e^{(T)}
\label{eq:target_conditioned_statistics}
$$
where $n_x$ is the number of rollout episodes with target location $x$, where $x\in\{\ell,c,r\}$. The symbols $\ell$, $c$, and $r$ denote the left, center, and right apple positions, respectively.
The directional component that remains across the left and right target conditions is defined as
\begin{equation}
H_{\mathrm{prior}}^{(T)} = \frac{\mu_{\ell}^{(T)}+\mu_{r}^{(T)}}{2}.
\label{eq:target_controlled_prior}
\end{equation}
$H_{\mathrm{prior}}^{(T)}$ captures the residual directional bias that remains after averaging across the left and right target conditions. Its sign follows the convention of $H^{(T)}$, with positive and negative values indicating residual right- and left-hand biases, respectively.

Finally, the responsiveness of the policy to the target location $x$ is defined as
\begin{equation}
H_{\mathrm{target}}^{(T)}=\frac{\mu_{r}^{(T)}-\mu_{\ell}^{(T)}
}{2}.
\label{eq:target_responsiveness}
\end{equation}
A positive $H_{\mathrm{target}}^{(T)}$ indicates that the HandPriorScore shifts toward the right robot side as the target moves from the left to the right. A larger value therefore indicates that the policy more strongly adjusts its initial hand-action direction according to the target location. The closer $H_{\mathrm{target}}^{(T)}$ is to zero, the weaker the influence of target-location cues on the policy's initial hand preference. The center-condition mean $\mu_c^{(T)}$ is not directly used to compute these two components. 
% and is instead reported as a supplementary statistic describing the policy behavior under a neutral target location.

\FloatBarrier
%%%%%%%%%%%%%%%%%%%%%%%%%%%%%%%%%%%%%%%%%%%%%%%%%%%%%%%%%%%%%%%%%%%%%%%%%%%%%%%%

\section{Experiment}

% 배치 때문에 미리 선언
\setcounter{figure}{1}

\begin{figure*}[t]
    \centering
    \includegraphics[width=\textwidth]{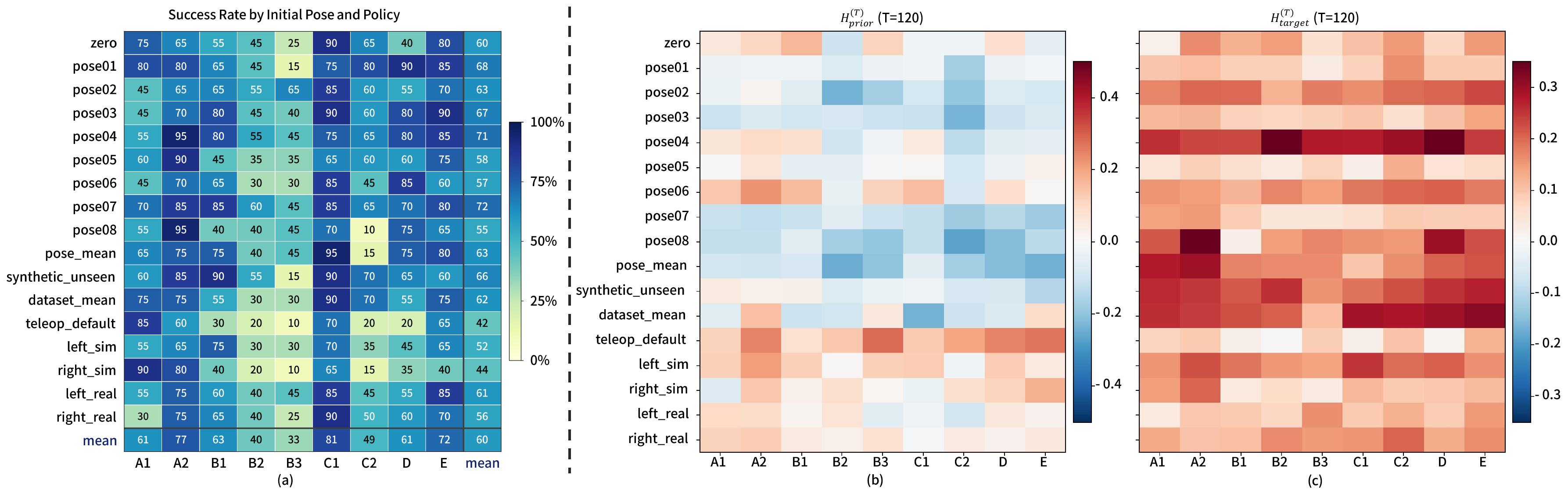}
    \caption{\textbf{Task success of initial-pose and policy dependence (left) and early hand preference during the initial decision window (right).} (a) PickApple success rate for each initial-pose–-policy pair. (b) Residual directional bias $H_{\mathrm{prior}}^{(T)}$ after averaging across the left and right target conditions. (c) Target-location responsiveness $H_{\mathrm{target}}^{(T)}$. The hand-prior metrics are computed using an initial decision window of T=120. Rows denote initial robot pose and columns denote policy variants; the last row and column in (a) report the corresponding averages.
    }
    \label{handpriorscore}
   
\end{figure*}
\setcounter{figure}{0}

\subsection{Training Datasets}

We select PickApple as the target task because it requires target-dependent hand selection, making it suitable for evaluating initial-pose dependence. The training datasets used in this study are summarized in Table~\ref{tab:dataset_types}.

\begin{table}[H]
\caption{Training Datasets and Observation Configurations}
\label{tab:dataset_types}
\centering
\scriptsize
\setlength{\tabcolsep}{3pt}
\renewcommand{\arraystretch}{1.15}

\resizebox{\columnwidth}{!}{%
\begin{tabular}{c | c c c}
\toprule
\textbf{Task ID}
& \textbf{Task Type}
& \textbf{Number of Tasks}
& \textbf{Camera} \\
\midrule

$P_{\mathrm{sim}}$
& \makecell[c]{PickApple Task \\ (Sim Teleop Data)}
& 1
& High and Wrist Cameras \\

\midrule

$P_{\mathrm{real}}$
& \makecell[c]{PickApple Task \\ (Real Teleop Data)}
& 1
& High Camera Only \\

\midrule

7-task
& \makecell[c]{Pick-Type Tasks \\ (Real Teleop Data)}
& 7
& High Camera Only \\

\midrule

13-common
& \makecell[c]{Pick and Non-Pick Tasks \\ (Real Teleop Data)}
& 13
& High Camera Only \\

\midrule

13-all
& \makecell[c]{Pick and Non-Pick Tasks \\ (Real Teleop Data)}
& 13
& \makecell[c]{High Camera Only (7) \\
               High and Wrist Cameras (6)} \\

\bottomrule
\end{tabular}%
}

\end{table}

They are broadly divided into the simulation dataset $P_{\mathrm{sim}}$  and several real-robot task groups from the Unitree teleoperation datasets. The $P_{\mathrm{sim}}$ dataset consists of simulated PickApple demonstrations collected from eight distinct initial robot poses shown in Fig.~\ref{poses} (right). For each initial pose, the apple was placed on either the left or right side of the robot, and 19 episodes were collected for each pose-–location combination. This resulted in a total of 304 episodes. The dataset size was chosen to remain comparable to a previous target-coupled two-pose configuration containing 306 episodes in Fig.~\ref{poses} (left). 
% By distributing a similar number of episodes across eight initial poses, we reduce the confounding effect of overall dataset size when comparing limited and diverse initial-pose coverage. - C에서 설명

The simulated task scene and robot configuration were designed to correspond to those of the real-robot PickApple dataset $P_{\mathrm{real}}$. However, $P_{\mathrm{sim}}$ contains observations from both the high and wrist cameras, whereas $P_{\mathrm{real}}$ contains only high-camera observations. The remaining dataset groups were constructed from the 13 real-robot manipulation datasets in the Unitree G1-Dex3 collection and organized according to task type and camera configuration \cite{unitree_dex3_datasets}. The 7-task group contains seven pick-type tasks, all of which use only high-camera observations. The 13-common group extends 7-task with six non-pick tasks. Although the original data for these six tasks include wrist-camera observations, the wrist-camera inputs are excluded in this configuration so that all 13 tasks share a common high-camera-only observation format. The 13-all group contains the same tasks as 13-common, but retains all available camera observations. Consequently, the seven pick-type tasks use only the high camera, whereas the six non-pick tasks use both the high and wrist cameras.

\subsection{Training Configuration}
%The datasets defined in Table~\ref{tab:dataset_types} are combined into the training configurations summarized in Table~\ref{tab:training_configuration}. These configurations are designed to compare the effects of real-data mixing, wrist-camera availability, and simulation sampling ratio on initial-pose dependence. All policies were fine-tuned from the pretrained GR00T N1.5-3B checkpoint \cite{GR00T}, with the training duration adjusted according to the size of each dataset mixture.
\begin{table}[thb]
\caption{Training configurations used in the comparative experiments.}
\label{tab:training_configuration}
\centering
\scriptsize
\setlength{\tabcolsep}{3pt}
\renewcommand{\arraystretch}{1.15}

\resizebox{\columnwidth}{!}{%
\begin{tabular}{c | c c c}
\toprule
\textbf{ID}
& \textbf{Training Data}
& \textbf{Camera}
& \textbf{Sim Ratio (\%)} \\
\midrule

A1
& $P_{\mathrm{sim}}$
& H
& 100.0 \\

A2
& $P_{\mathrm{sim}}$
& W
& 100.0 \\

\midrule

B1
& $P_{\mathrm{sim}} + P_{\mathrm{real}}$
& H
& N(45)  \\

B2
& $P_{\mathrm{sim}} + \text{7-task}$
& H
& N(9.4) \\

B3
& $P_{\mathrm{sim}} + \text{13-common}$
& H
& N(4.6) \\

\midrule

C1
& $P_{\mathrm{sim}} + P_{\mathrm{real}}$
& \makecell[c]{
    $P_{\mathrm{sim}}$: W, 
    $P_{\mathrm{real}}$: Z
}
& N(45) \\

C2
& $P_{\mathrm{sim}} + \text{13-all}$
& \makecell[c]{
    $P_{\mathrm{sim}}$: W,  7-task: Z, 
    non-pick: W
}
& N(4.6) \\

\midrule

D
& $P_{\mathrm{sim}} + \text{7-task}$
& H
& $\alpha$(45.5) \\

\midrule

E
& $P_{\mathrm{sim}} + \text{7-task}$
& \makecell[c]{
    $P_{\mathrm{sim}}$: W,
    7-task: Z
}
& $\alpha$(45.5) \\

\bottomrule
\end{tabular}%
}

\vspace{2pt}
\begin{minipage}{\columnwidth}
\scriptsize
\textit{Camera:}
H = high camera only;
W = high and wrist cameras;
Z = high camera with zero-filled wrist streams.
\textit{Sim ratio:}
N = natural ratio determined by dataset size;
$\alpha$ = reweighted sampling ratio.
\end{minipage}

\end{table}

Table~\ref{tab:training_configuration} isolates four factors. The A group uses only the single-simulation-task $P_\mathrm{sim}$ dataset. A1 and A2 compare high-camera-only and wrist-enabled training. The B group combines $P_\mathrm{sim}$ with increasingly broad real-robot datasets under a common high-camera-only observation format. No sampling reweighting is applied; the simulation ratio is naturally determined by the relative sizes of the constituent datasets. The C group uses both high- and wrist-camera streams. Actual wrist observations are used where available, and missing streams are zero-filled to maintain consistent input dimensionality. Both D and E are trained on $P_\mathrm{sim}$+7-task, with the simulation sampling ratio reweighted to approximately 45.5\%. This value is selected to be comparable to the approximately 45\% natural simulation ratio of B1 and C1, which are trained on $P_\mathrm{sim}$+$P_\mathrm{real}$. The 45.5\% ratio is used as a controlled comparison point and is not intended to represent an optimal sampling ratio. Identifying the optimal balance between target-task and auxiliary data remains an open problem. D and E differ only in camera format. All policies were fine-tuned from GR00T-N1.5-3B \cite{GR00T_N15, unitree_lerobot}.

\subsection{Initial-Pose Set and Evaluation Protocol}

\begin{figure}[H]
  \centering
  \includegraphics[width=\columnwidth]{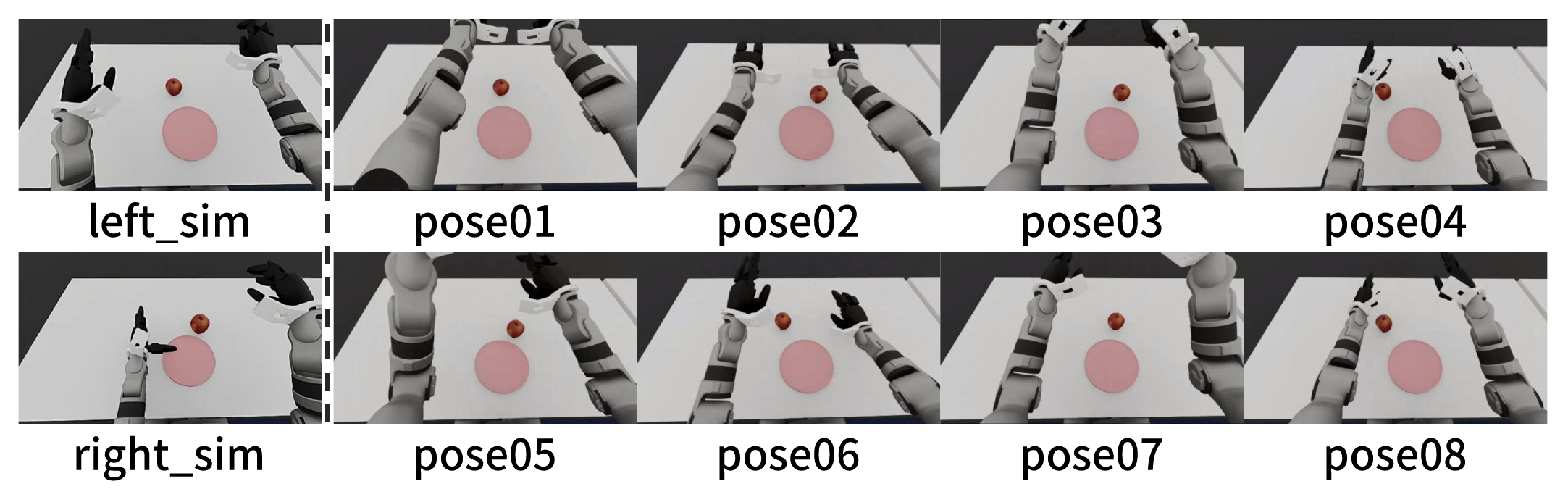}
  \caption{\textbf{Simulation initial-pose sets used before and after pose diversification.}  Left: two target-conditioned initial poses used in the limited-pose setting. Right: eight diversified initial poses varying arm height, wrist orientation, and bilateral arm symmetry.}
  \label{poses}
  
\end{figure}

\setcounter{figure}{2}

\begin{figure}[tbh]
    \centering
    \includegraphics[width=\columnwidth]{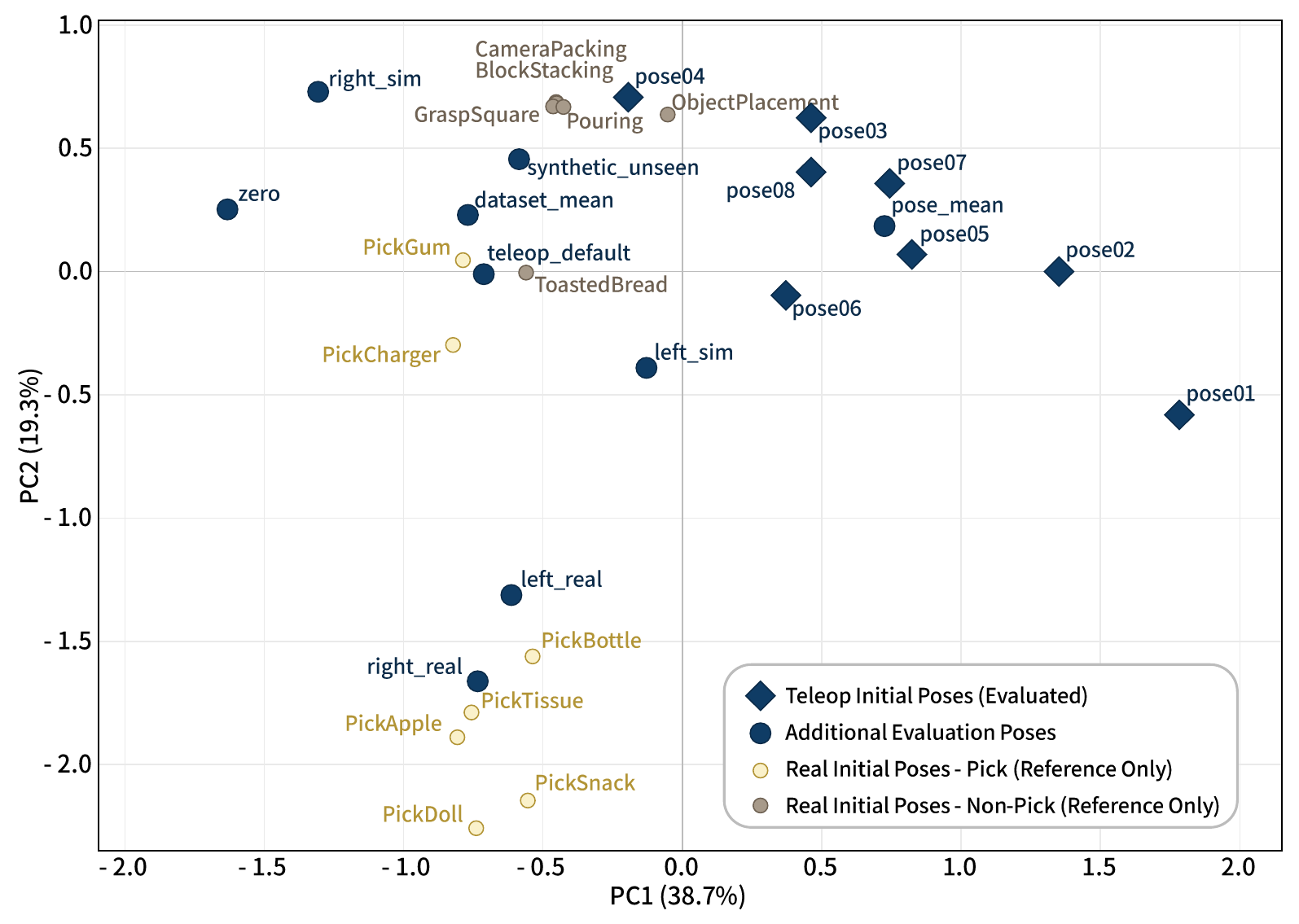}
    \caption{\textbf{PCA projection of evaluation initial poses and task-level distributions.} Blue diamonds denote 8 diversified PickApple training poses, while blue circles denote the remaining evaluation initial configurations. Yellow and gray circles respectively represent the task-level initial-state means of the pick-type and non-pick real-robot datasets.
    %The first two principal components explain 38.7\% and 19.3\% of the total variance. 
    %The projection visualizes the relative coverage and proximity between the evaluated initial poses and the initial-state distributions represented in the training datasets.
    }
    
    \label{init pose distribution}
\end{figure}

The limited setting used before pose diversification consisted of the two initial configurations shown in Fig.~\ref{poses} (left). The upper pose (left\_sim) was used when the apple was placed on the left or at the center, while the lower pose (right\_sim) was used for the right or center target conditions. Consequently, the initial robot pose was coupled with the target location in this setting. Preliminary rollouts showed that the policy tended to use the initial robot state as a cue for hand selection rather than responding consistently to the target observation. This observation motivated the construction of the diversified eight-pose set shown in Fig.~\ref{poses} (right), with the aim of reducing the coupling between the initial pose and target location while broadening the pose coverage. Unlike the limited two-pose setting, each diversified pose was paired with both left- and right-side apple placements during the collection of $P_\mathrm{sim}$, rather than being assigned to a particular target side. Their locations in the initial-state distribution are shown as blue diamonds in Fig.\ref{init pose distribution}.
The remaining evaluation-pose names indicate how each configuration was constructed. pose\_mean is the joint-wise mean of pose01--pose08, and zero is the all-zero joint configuration. teleop\_default is an initial configuration selected from one right-target demonstration in the pre-diversification dataset; right\_sim is the joint-wise mean across all right-target initial configurations in that dataset. left\_real and right\_real are the first-frame configurations of selected left- and right-target demonstrations from $P_\mathrm{real}$, respectively.
%they are distinct from the task-level PickApple mean shown in yellow in Fig.~\ref{init pose distribution}. 
dataset\_mean is a sim-neutral candidate selected from configurations derived from the mean of the pre-diversification dataset. Finally, synthetic\_unseen is an out-of-range drift pose constructed by extrapolating the 14 arm dimensions of pose\_mean along a principal direction in PCA space. 
%/while keeping its 14 hand dimensions unchanged.

The policies listed in Table~\ref{tab:training_configuration} were subsequently evaluated through closed-loop rollouts from 17 initial poses. Each initial-pose-policy condition was evaluated over 20 rollout episodes, and task success was manually assessed. A rollout was considered successful if either hand grasped the apple, transported it to the pink target plate, released it, and completed the subsequent arm-lift motion; failure at any stage was counted as unsuccessful. 

%Success was independent of whether the selected hand matched the target side and therefore measures overall PickApple task completion, while target-dependent hand selection was evaluated separately using $H_{\mathrm{target}}^{(T)}$.

%%%%%%%%%%%%%%%%%%%%%%%%%%%%%%%%%%%%%%%%%%%%%%%%%%%%%%%%%%%%%%%%%%%%%%%%%%%%%%%%

\section{Results}

\subsection{Initial-Pose Dependence Produces Policy-Specific Success Variations}

Fig.~\ref{handpriorscore}(a) shows substantial success variation across both initial poses and policies. For example, A1 achieves relatively high success across many initial poses but drops to 30\% at right\_real, and C2 ranges from 10\% to 80\% across poses. The same initial pose also produces markedly different outcomes across policies; at pose\_mean, success ranges from 15\% to 95\%. Similar policy-dependent differences appear at teleop\_default and right\_sim. These results demonstrate an initial-pose--policy interaction in which task success depends jointly on the initial configuration and the learned policy. We next examine whether this interaction is reflected in early hand-selection behavior.

\subsection{HandPriorScore Reveals Residual Hand Bias and Target Responsiveness}

Fig.~\ref{handpriorscore}(b) shows the residual left--right hand bias, defined in Eq.~\eqref{eq:target_controlled_prior}, after averaging across target-side conditions. Fig.~\ref{handpriorscore}(c) shows target responsiveness, defined in Eq.~\eqref{eq:target_responsiveness}, which measures how strongly the initial hand preference shifts as the apple moves from the left to the right. Notably, the teleop\_default pose exhibits a strong residual right-hand bias and weak target-location responsiveness. This pattern suggests a pose-conditioned shortcut in which the initial robot state outweighs the target-side cue during early hand selection. It does not imply that visual observations are ignored; their influence is examined through the modality interventions below.

\subsection{Pose Diversification and Targeted Augmentation Improve Robustness}

\begin{figure}[H]
  \centering
  \includegraphics[width=\columnwidth]{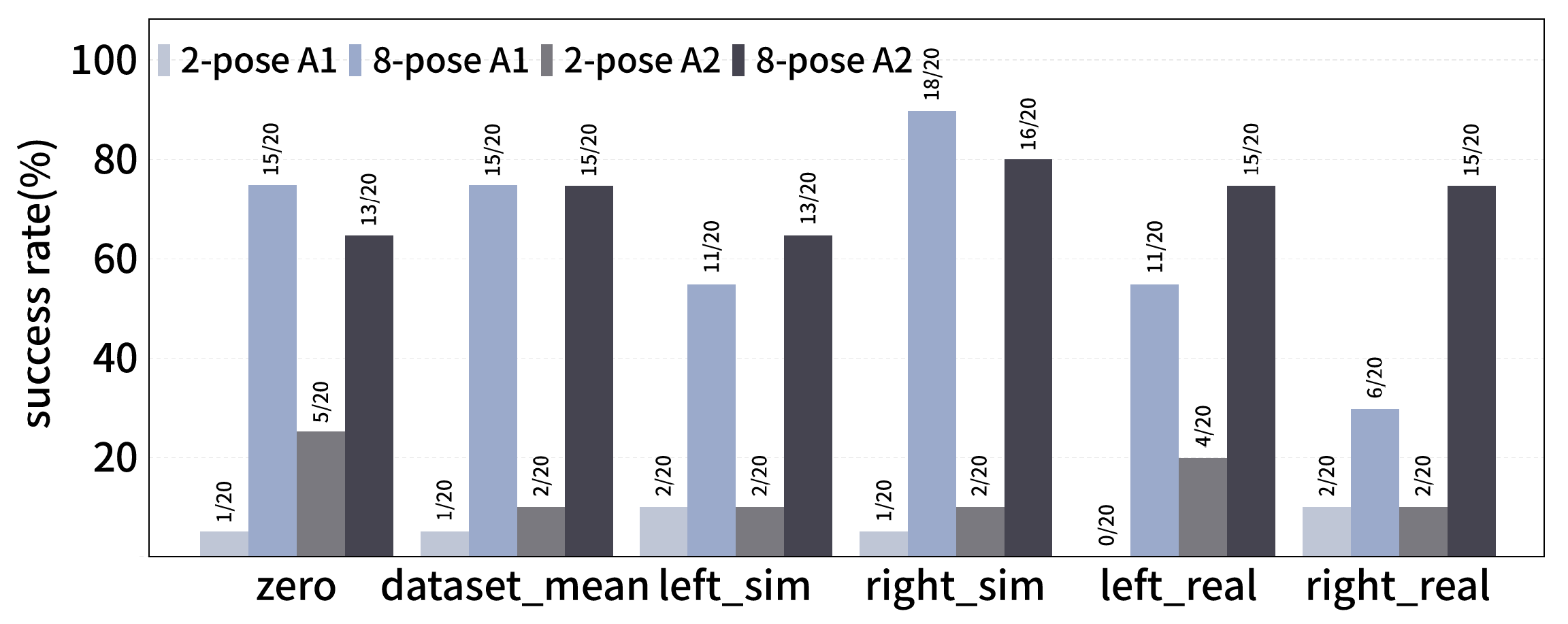}
  \caption{\textbf{Effect of initial-pose diversification on PickApple success.} A1 and A2 were trained using either the pre-diversification dataset collected from two target-coupled initial poses or the diversified dataset collected from eight initial poses. }
  \label{pose_diversification}
\end{figure}

As shown in Fig.~\ref{pose_diversification}, pose diversification increased the mean success rate from 5.8\% to 63.3\% for A1 and from 14.2\% to 72.5\% for A2 across six evaluation poses. These gains under both camera settings indicate that broader initial-state coverage reduces pose-dependent failures.

 \begin{figure}[!thbp]
      \centering
      \includegraphics[width=\columnwidth]{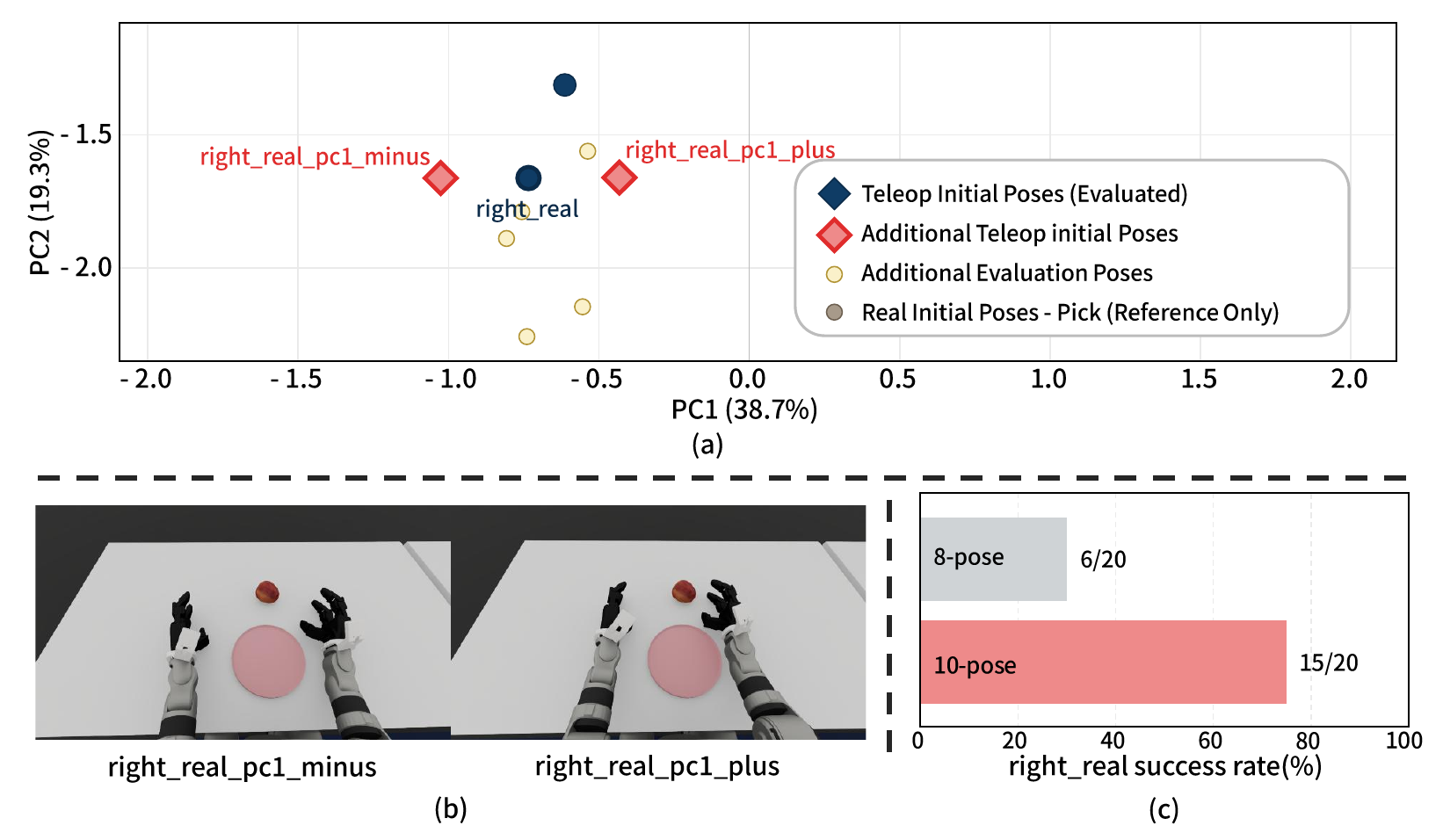}
      \caption{\textbf{Targeted pose augmentation around the right\_real configuration.} (a) Enlarged PCA projection of the local initial-state region. The two red diamonds denote additional collection poses placed on either side of right\_real along PC1. (b) Robot configurations images corresponding to right\_real\_pc1\_minus and right\_real\_pc1\_plus. (c) Success rate at right\_real for A1 trained with the original eight-pose dataset and the augmented ten-pose dataset.}
      \label{targeted_pose_augmentation}
   \end{figure}

Although A1 achieved an average success rate of 61.5\%, its success at right\_real was only 30\%, indicating insufficient local pose coverage. We therefore added two collection poses on either side of right\_real along PC1, expanding the dataset from eight to ten poses. As shown in Fig.~\ref{targeted_pose_augmentation}, this targeted augmentation increased success at right\_real from 30\% to 75\%, demonstrating that local coverage of failure-prone regions can improve pose robustness.

\subsection{Effects of Auxiliary Data, Wrist Observations, and Simulation Ratio}

Unless otherwise stated, overall success is averaged over 17 initial poses with 20 rollouts per pose, totaling 340 rollouts per policy. We additionally report success over the eight diversified poses and analyze diagnostic reference poses separately to preserve localized failure patterns.

\textbf{Effect of wrist observations.}
A1 and A2 differed only in wrist-camera availability. Adding wrist observations increased overall success from 61.5\% to 76.8\% and diversified-pose success from 56.9\% to 81.2\%, with gains on seven of eight poses. Over the diversified poses, $H_{\mathrm{prior}}^{(T)}$ remained near neutral and $H_{\mathrm{target}}^{(T)}$ increased slightly. At teleop\_default, however, wrist observations strengthened the right-hand bias, increasing $H_{\mathrm{prior}}^{(T)}$ from $+0.12$ to $+0.25$, raising wrong-hand crossing from 6\% to 33\%, and reducing success from 85\% to 60\%. Thus, wrist observations improved overall robustness but did not consistently reduce pose-conditioned hand bias.

\textbf{Effect of auxiliary real data under natural sampling.}
A1, B1, B2, and B3 use the same high-camera-only format but include different real-robot datasets. Adding $P_{\mathrm{real}}$ slightly increased overall success from 61.5\% to 62.9\%, whereas adding 7-task and 13-common reduced it to 40.3\% and 32.6\%, respectively. The corresponding diversified-pose means were 56.9\%, 65.6\%, 45.6\%, and 40.0\%. Performance at teleop\_default and right\_sim also declined as the auxiliary dataset grew and the natural simulation ratio decreased. The resulting failures differed across policies: B2 showed increased bimanual indecision, while B3 exhibited the strongest residual right-hand prior at teleop\_default. These results therefore reflect both auxiliary-data composition and reduced exposure to $P_{\mathrm{sim}}$.

\textbf{Effect of simulation-ratio reweighting.}
Compared with B2, D increases the simulation ratio from 9.4\% to 45.5\% while keeping the training data and camera format fixed. Reweighting improved overall success from 40.3\% to 61.5\% and diversified-pose success from 45.6\% to 74.4\%, while eliminating bimanual indecision over the diversified poses and increasing $H_{\mathrm{target}}^{(T)}$ from 0.13 to 0.18. At teleop\_default, however, success remained at 20\% as bimanual indecision decreased and wrong-hand crossing increased from 0\% to 58\%. Thus, greater $P_{\mathrm{sim}}$ exposure improved robustness on diversified poses but did not resolve the diagnostic failure.

\textbf{Effect of mixed wrist availability.}
With the same data mixture and simulation ratio, C1 uses wrist images for $P_{\mathrm{sim}}$ and zero-filled wrist streams for $P_{\mathrm{real}}$, while B1 is high-camera-only. This increased overall success from 62.9\% to 80.9\% and diversified-pose success from 65.6\% to 78.8\%, while improving teleop\_default success from 30\% to 70\%. At the higher simulation ratio, E also outperformed D overall (72.4\% versus 61.5\%), although their diversified-pose means were similar. The difference was concentrated at diagnostic reference poses, showing that similar diversified-pose averages can conceal substantial differences in reference-pose robustness and hand-selection behavior.

\subsection{Modality and Joint-Level Interventions Localize the teleop\_default Failure}

The teleop\_default pose was selected as a diagnostic case because it exhibited a strong residual right-hand prior, weak target responsiveness, and frequent right-hand attempts on left-target episodes. We first tested the influence of wrist-camera observations and then localized the remaining sensitivity through joint-level interventions.

   \begin{figure}[bht]
      \centering      
      \includegraphics[width=\columnwidth]{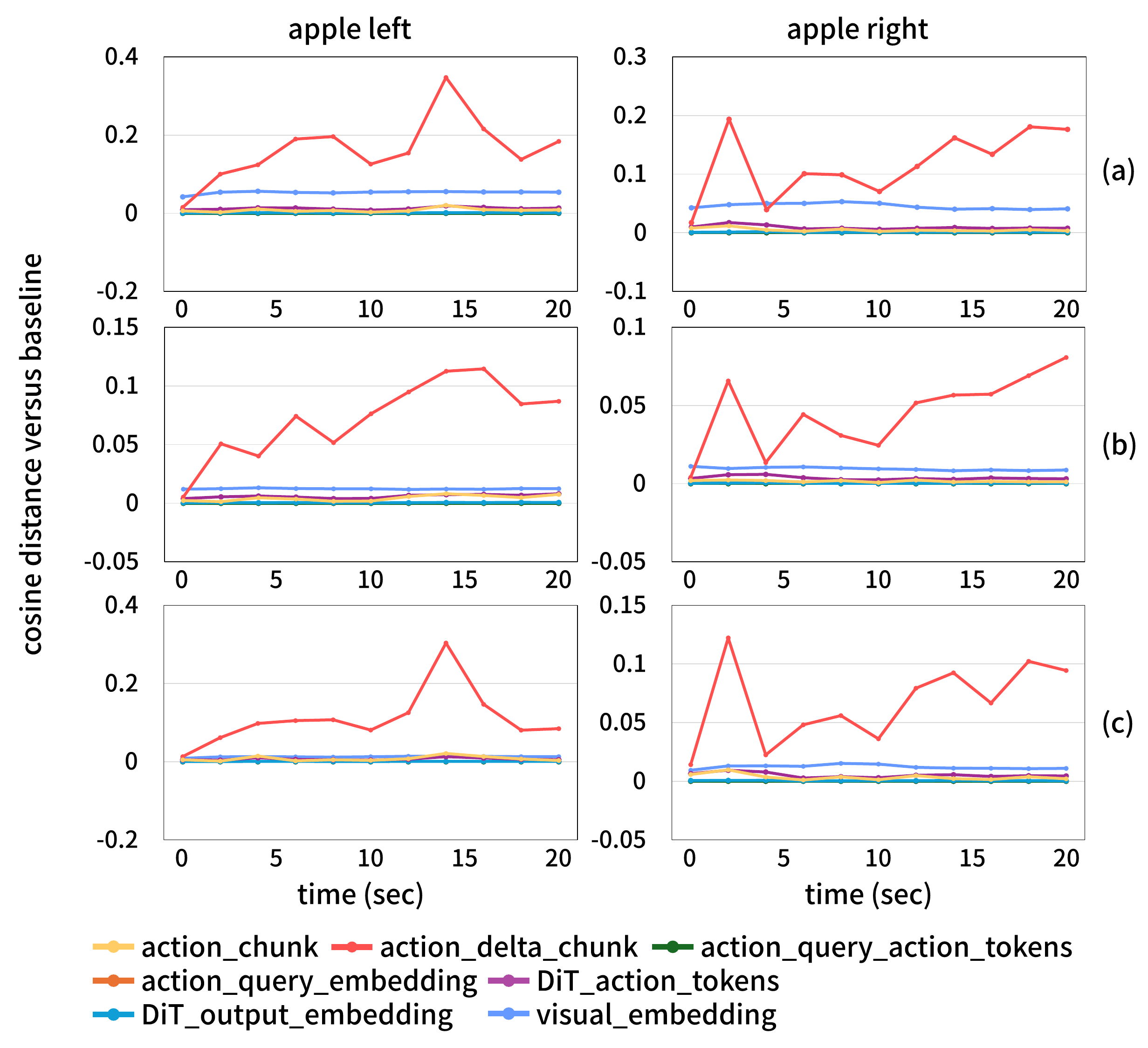}
      \caption{Wrist-mask perturbation for A2 at teleop\_default. (a) Both-wrist camera masking, (b) left-wrist camera masking, and (c) right-wrist camera masking. Columns correspond to left- and right-side apple placements. Each curve reports the cosine distance from the paired full-observation baseline for the visual, intermediate-token, and predicted-action representations over 0--20~s.}
      \label{wrist mask token analysis}
   \end{figure}
   
\textbf{Wrist-camera intervention.} We compared the full-observation baseline with left-, right-, and both-wrist-masked conditions. Across four evaluation runs, baseline, right-wrist-mask, and left-wrist-mask achieved similar success rates, while masking both wrists reduced success to 20.0\%. Despite their similar overall success rates, on left-target episodes, right-wrist masking reduced the right-hand attempt rate from 97.1\% to 40.5\%, whereas left-wrist masking yielded 88.2\%. Thus, wrist observations support task performance, and the right-wrist stream strengthens the right-hand preference at teleop\_default, but does not fully explain the wrong-hand behavior.

The token analysis in Fig.~\ref{wrist mask token analysis} shows that wrist masking produced only small changes in the absolute action chunk (cosine distance: 0.002--0.008), but larger changes in the state-relative action-delta chunk defined in Eq.~\eqref{eq:relative_action} (0.04--0.16). The effect was generally strongest for both-wrist masking, followed by right- and left-wrist masking. This suggests that wrist observations influence incremental motion planning and approach or grasp adjustment, even when absolute joint targets remain anchored to the current state. Because right-wrist masking reduced but did not eliminate wrong-hand attempts, we next tested whether the remaining sensitivity could be localized to the initial arm configuration.

\textbf{Joint-level intervention.} We constructed counterfactual initial configurations by swapping selected right-arm joint values between teleop\_default and right\_sim while keeping all remaining joints fixed.

   \begin{figure}[htb]
      \centering
      \includegraphics[width=\columnwidth]{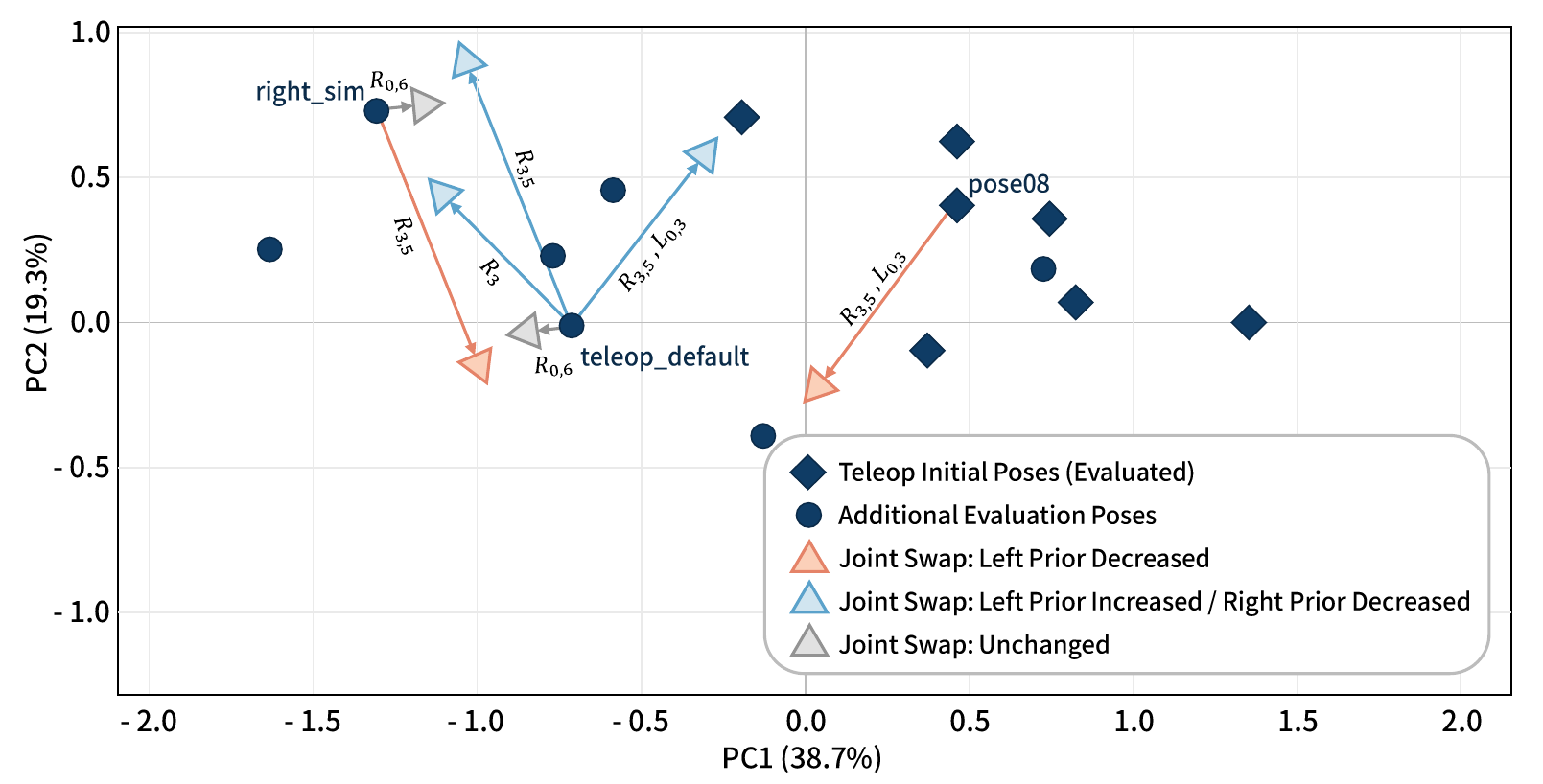}
      
      \caption{\textbf{Joint-level counterfactual interventions.} The A2 intervention between teleop\_default and right\_sim swaps $R_{\mathrm{arm},3}$ and $R_{\mathrm{arm},5}$, while the control swaps $R_{\mathrm{arm},0}$ and $R_{\mathrm{arm},6}$. The broader pose08--teleop\_default intervention swaps the four arm joints with the largest absolute position differences. Arrow labels indicate the swapped arm and joint indices using zero-based indexing.}
      
      \label{joint intervention}
   \end{figure}

Although teleop\_default and right\_sim have broadly similar initial arm configurations, they produce markedly different asymmetric hand preferences. A joint-wise comparison showed that the absolute position differences at $R_{\mathrm{arm},3}$ and $R_{\mathrm{arm},5}$ were 1.269 and 1.389, respectively, whereas all other right-arm joints differed by less than 0.39. We therefore selected $R_{\mathrm{arm},3}$ and $R_{\mathrm{arm},5}$ for the main intervention. The less-divergent joints $R_{\mathrm{arm},0}$ and $R_{\mathrm{arm},6}$ were used as a two-joint control to test whether the behavior changed merely because two right-arm joints were perturbed.
As shown in Table~\ref{joint_intervention_c4}, swapping $R_{\mathrm{arm},3}$ and $R_{\mathrm{arm},5}$ from right\_sim into teleop\_default reduced the right-hand attempt rate from 7/8 to 0/4, while swapping only $R_{\mathrm{arm},3}$ yielded 3/10 and the control swap remained high at 9/9. The reverse swap increased the rate at right\_sim from 0/12 to 7/7, whereas the control remained at 0/12. Thus, the two most-divergent joints suppressed or induced the asymmetric right-hand preference, unlike the lower-difference control joints.

\begin{table}[thb]
\caption{Joint-level intervention results for A2. Values denote right-hand attempt rates on left-target episodes.}

\label{joint_intervention_c4}
\centering
\scriptsize
\setlength{\tabcolsep}{4pt}
\renewcommand{\arraystretch}{1.12}
\begin{tabular}{l l c}
\toprule
\textbf{Base pose}
& \textbf{Intervention}
& \textbf{Right-hand rate} \\
\midrule

teleop\_default
& None
& $7/8$ (87.5\%) \\

teleop\_default
& Swap $R_{\mathrm{arm},3}$
& $3/10$ (30.0\%) \\

teleop\_default
& Swap $R_{\mathrm{arm},3}$ and $R_{\mathrm{arm},5}$
& $0/4$ (0.0\%) \\

teleop\_default
& Swap $R_{\mathrm{arm},0}$, $R_{\mathrm{arm},6}$ (control)
& $9/9$ (100\%) \\

\midrule

right\_sim
& None
& $0/12$ (0.0\%) \\

right\_sim
& Swap teleop $R_{\mathrm{arm},3}$ and $R_{\mathrm{arm},5}$
& $7/7$ (100\%) \\

right\_sim
& Swap teleop $R_{\mathrm{arm},0}$, $R_{\mathrm{arm},6}$ (control)
& $0/12$ (0.0\%) \\

\bottomrule
\end{tabular}
\end{table}

We further tested whether the prior-removal effect transferred across policies using the broader pose08--teleop arm intervention. 
For this intervention, swapping only $R_{\mathrm{arm},3}$ and $R_{\mathrm{arm},5}$ produced a physically infeasible configuration, so we swapped the initial joint-position values with the largest absolute position differences: $R_{\mathrm{arm},3}$ and $R_{\mathrm{arm},5}$ and $L_{\mathrm{arm},0}$ and $L_{\mathrm{arm},6}$.
As shown in Table~\ref{joint_intervention_generalization}, swapping these four joint values from teleop\_default into pose08 reduced left-hand attempts on right-target episodes from 4/14 to 0/12 for B1 and from 11/13 to 1/9 for C2. The reverse swap was less consistent, producing no change for B1 and an increase from 2/7 to 4/10 for C2. Thus, the prior-removal effect transferred across both policies, whereas reverse induction remained policy dependent.

\begin{table}[tbh]
\caption{Cross-policy generalization of the four-joint arm intervention.
Values denote left-hand attempt
rates on right-target episodes.}
\label{joint_intervention_generalization}
\centering
\scriptsize
\setlength{\tabcolsep}{3.2pt}
\renewcommand{\arraystretch}{1.12}

\resizebox{\columnwidth}{!}{%
\begin{tabular}{c l l c c}
\toprule
\textbf{Policy}
& \textbf{Base pose}
& \textbf{Source pose}
& \textbf{Original}
& \textbf{After swap} \\
\midrule

B1
& pose08
& teleop\_default
& $4/14$ (28.6\%)
& $0/12$ (0.0\%) \\

C2
& pose08
& teleop\_default
& $11/13$ (84.6\%)
& $1/9$ (11.1\%) \\

\midrule

B1
& teleop\_default
& pose08
& $0/11$ (0.0\%)
& $0/12$ (0.0\%) \\

C2
& teleop\_default
& pose08
& $2/7$ (28.6\%)
& $4/10$ (40.0\%) \\

\bottomrule
\end{tabular}%
}
\end{table}

Together, these results show that wrist vision modulates hand preference and task performance but does not fully explain the teleop\_default failure. In A2, bidirectional swaps of $R_{\mathrm{arm},3}$ and $R_{\mathrm{arm},5}$ suppressed or induced the right-hand preference, unlike the lower-difference control swaps. The broader four-joint intervention transferred prior removal across B1 and C2, although reverse induction remained policy dependent. These findings identify the localized initial arm configuration as a direct causal handle on the pose-conditioned hand prior, while wrist observations modulate its behavioral expression.

\FloatBarrier
%%%%%%%%%%%%%%%%%%%%%%%%%%%%%%%%%%%%%%%%%%%%%%%%%%%%%%%%%%%%%%%%%%%%%%%%%%%%%%%%

\section{CONCLUSIONS}

This work characterized initial-pose dependence in GR00T-based humanoid dual-arm manipulation as a policy-induced hand prior, quantified by residual hand bias and target responsiveness. Evaluations across initial poses and policies revealed strong initial-pose--policy interactions, including a persistent asymmetric hand preference with weak target responsiveness at a particular initial pose. Wrist-camera interventions showed that wrist observations influence hand preference and task performance but do not fully explain the wrong-hand behavior. Bidirectional swaps of the two most-divergent joints either suppressed or induced the preference, but swaps of less-divergent control joints did not, identifying the localized initial arm configuration as a causal handle on the pose-conditioned prior.

Training experiments further showed that robustness depends on initial-pose coverage and data composition. Expanding a target-coupled two-pose dataset to eight diversified poses substantially improved performance, while targeted augmentation increased the success rate from 30\% to 75\% at a previously low-performing initial pose. Wrist observations were generally beneficial for task performance, but adding real or auxiliary data was not consistently effective when it reduced target-task exposure or introduced mismatched observation availability. These results highlight the importance of pairing each initial pose with multiple target locations, covering failure-prone initial-state regions, and preserving sufficient target-task exposure in mixed training datasets.

This study is limited to a single task, robot embodiment, and policy family, with small and unequal sample counts in several intervention conditions. Future work will examine whether these findings transfer across tasks and VLA architectures.


%%%%%%%%%%%%%%%%%%%%%%%%%%%%%%%%%%%%%%%%%%%%%%%%%%%%%%%%%%%%%%%%%%%%%%%%%%%%%%%%

\bibliographystyle{IEEEtran}
\bibliography{root}

\end{document}